\documentclass[letterpaper, 10 pt, conference]{ieeeconf}  

\IEEEoverridecommandlockouts                              

\usepackage{graphics} 
\usepackage{epsfig} 
\usepackage{mathptmx} 
\usepackage{times} 
\usepackage{amsmath} 
\usepackage{amssymb}  
\usepackage{xcolor}

\usepackage{caption}
\usepackage{subcaption}
\usepackage{multirow}
\usepackage{multicol}

\title{\LARGE \bf
DCAD: Decentralized Collision Avoidance with Dynamics Constraints for Agile Quadrotor Swarms
}

\author{Senthil Hariharan Arul$^{1}$ and Dinesh Manocha$^{2}$
\thanks{$^{1}$Senthil Hariharan Arul is with Department of Electrical and Computer Engineering, University of Maryland, College Park, USA        {\tt\small sarul1@umd.edu}}%
\thanks{$^{2}$Dinesh Manocha is with the Department of Computer Science, Electrical and Computer Engineering, University of Maryland, College Park, USA
        {\tt\small dm@cs.umd.edu}}%
}

\begin{document}

\maketitle
\thispagestyle{empty}
\pagestyle{empty}
\begin{abstract}
We present a  novel, decentralized collision avoidance algorithm for navigating a swarm of quadrotors in dense environments populated with static and dynamic obstacles. Our algorithm relies on the concept of Optimal Reciprocal Collision Avoidance (ORCA) 
and utilizes a flatness-based Model Predictive Control (MPC) to generate local collision-free trajectories for each quadrotor. 
We feedforward linearize the non-linear dynamics of the quadrotor and subsequently use this linearized model in our MPC framework. 
Our method approach tends to compute safe trajectories that avoid quadrotors from entering each other's downwash regions during close proximity maneuvers. In addition, we account for the uncertainty in the position and velocity sensor data using Kalman filter. We evaluate the performance of our algorithm with other state-of-the-art decentralized methods and demonstrate its superior performance in terms of smoothness of generated trajectories and lower probability of collision during high velocity maneuvers. 
\end{abstract}
\section{Introduction}
Due to their agility, ease of deployment, decreasing costs, and small size, quadrotors are extensively used in robotics and related areas. Their 3-D navigation capabilities are favorable for tasks such as surveillance, target tracking, search \& rescue, etc. Many applications use a coordinating swarm of quadrotors for efficiency, where large 3D regions can be simultaneously covered by different quadrotors. Agile, high-velocity quadrotor navigation is especially important in applications involving disaster response, where time is critical for effective rescue operations \cite{Agile}. Further, for safe implementation, it is important to avoid collisions with the obstacles in the scene as well as other quadrotor agents during the high-velocity maneuvers. Such applications require navigation algorithms that can rapidly adapt to the changes in the environment, account for sensor uncertainty, and scale to a large number of swarm agents.

Many centralized \cite{Kushleyev,Augugliaro,Chen,Preiss} and decentralized \cite{LQG,LQR} planning approaches have been proposed for collision-free navigation of quadrotors in a swarm. Centralized methods scale poorly and are generally not adaptable to dynamic environments but can provide guarantees on trajectory smoothness, optimality, etc. In contrast, decentralized methods are scalable and are adaptable to changes in the environment, but cannot provide such guarantees on the generated trajectories. 

{Quadrotor dynamics is non-linear due to the sinusoidal relationships required to describe its orientation \cite{quadmodel}.}  
Collision avoidance methods that account for  such non-linearities either do not run in real-time~\cite{Augugliaro,Chen} or use a computationally expensive controller such as N-MPC~\cite{Zhu}, thus limiting the available on-board computational power for other applications like perception and communication. In addition, an N-MPC solution is susceptible to converging to local minima \cite{Greeff}. 
In contrast, other methods ~\cite{LQG,LQR} reduce the complexity by linearizing the quadrotor dynamics about an equilibrium point (usually about the quadrotor's hover configuration). 
This linearized model is valid near the equilibrium point, but its performance decreases during aggressive (high-velocity) maneuvers, where large pitch and roll is required \cite{hoverlimitation}. 
Further, to account for downwash in the collision avoidance algorithm, the quadrotors are modeled as axis-aligned ellipsoids, disregarding the quadrotor orientation \cite{Preiss, LQR}, though the downwash region would vary in relation to the quadrotor orientation \cite{downwash2}. 
{

\noindent {\bf Main Results:} We present a novel decentralized realtime approach (DCAD) for navigation of large quadrotor swarms in dynamic environments. Our approach is general and makes no assumptions about the obstacles or the environment. To handle the non-linear dynamics constraints and enable fast maneuvers, we present two novel algorithms:
\begin{enumerate}
    \item An on-line collision avoidance algorithm for navigation in dynamic environments that accounts for quadrotor dynamics using flatness-based feedforward linearization and MPC. In contrast to linearizing about the hover point, we can still incorporate the non-linearities in the system using an inverse map. 
    \item An algorithm to incorporate downwash into the construction of collision avoidance constraints based on ORCA. In contrast to using axis-aligned ellipsoids to consider downwash, our algorithm incorporates quadrotor attitude by modeling neighboring pair of quadrotors as a combination of a sphere and an {\color{black}oriented ellipsoid}. 
\end{enumerate}
We combine these algorithms with ORCA constraints for collision avoidance~\cite{RVO} to compute the local trajectory for each quadrotor in a decentralized manner.
{\color{black}In addition, our method incorporates sensing uncertainty using Kalman filtering and provides scalability with respect to the number of agents.
}
In practice, our algorithm takes about {\color{black} 5 ms on average} to calculate a new collision-avoiding control input for an agent in the presence of {\color{black} 8} obstacles. Moreover, our DCAD results in better behaviour in terms of smoother trajectories and collision avoidance during high-velocity \textcolor{black}{maneuvers}, as compared to prior decentralized methods based on ORCA \cite{ORCA}, AVO \cite{AVO}, LQR-obstacles \cite{LQR}, and LSwarm \cite{LSwarm}.}

The rest of the paper is organized as follows. In Section II, we summarize state-of-the-art methods in collision avoidance and quadrotor control. In Section III, we introduce our quadrotor model and the notion of feedforward linearization. In Section IV, we present our decentralized collision avoidance algorithm that considers dynamics constraints. In Section V, we describe its implementation and highlight the benefits over prior methods. 
compared to other state-of-the-art methods. 
\section{Previous Work}
In this section, we give a brief overview of prior work in quadrotor control and collision avoidance. 

\subsection{Quadrotor Control}

In prior literature  \cite{LQG,LQR,Mellinger_Hover,Hoffmann}, quadrotor dynamics is handled by linearizing the system dynamics about the hover point to facilitate the use of a linear controller such as an LQR \cite{lqrcontrol} or a linear Model Predictive Control (MPC) \cite{mpccontrol}. These methods facilitate reduced computational overhead compared to non-linear controllers, thereby allowing more on-board processing power for other applications like perception or communication.
However, aggressive (high-velocity, high-acceleration) maneuvers require large attitude deviations from the hover point, and the performance is reduced when hover-point linearization is used \cite{linearvsnonlinear}. Kamel et al. \cite{Kamel} and Zhu et al. \cite{Zhu} present a non-linear Model Predictive Control-based (NMPC-based) collision avoidance method, that models the full non-linear quadrotor dynamics. 
The NMPC-based algorithm~\cite{Zhu} takes about 16ms to compute a collision avoiding control input for a quadrotor with 6 neighboring obstacles. 
Flatness-based feedforward controller for trajectory tracking is proposed in \cite{Ferrin,Greeff}. 
Unlike the methods that linearize about the hover point, these feedforward methods do not make small angle assumptions about the roll and pitch of the quadrotors. Controllers based on feedforward linearization and flatness have shown better performance in terms of computation time and, unlike NMPC, they are not sensitive to the choice of the initial trajectory or susceptible to local minima convergence issues \cite{Greeff}. Because of these advantages, we use flatness-based feedforward linearization to model the quadrotor dynamics in our algorithm. 


\subsection{Collision Avoidance}
Prior research can be grouped into two broad areas, centralized trajectory generation and reactive collision avoidance.

\subsubsection{Centralized Trajectory Generation}
Augugliaro et al. \cite{Augugliaro} and Chen et al. \cite{Chen} propose a centralized algorithm that relies on solving a sequential convex program to generate collision-free trajectories for a swarm. Kushleyev et al. \cite{Kushleyev} present a method that generates collision-free trajectories for a swarm of quadrotors by solving a Mixed Integer Quadratic Program (MIQP).  Due to the high computational overhead and centralized nature of MIQP, the algorithm scales exponentially with the number of agents. 
Preiss et al.~\cite{Preiss} reduce the computation cost by decomposing the collision-free trajectory generation problem into a discrete collision-free path planner and a trajectory optimizer. Hamer et al.~\cite{Hamer} present a parallel formulation for fast generation of collision-free trajectories in multi-agent scenarios. 
Centralized trajectory generation can guarantee optimality in terms of minimum path length, time to reach the goal, or fuel cost. However, these methods can be limited in terms of real-world urban scenarios due to the dynamic nature of the environment, a sudden change in mission, or covering very large areas. 



\subsubsection{Reactive Collision Avoidance}
Velocity Obstacle (VO) \cite{VO}-based methods such as RVO \cite{RVO} provide decentralized collision avoidance by locally altering the trajectories for agents with single-integrator dynamics. 
Constraints in RVO \cite{RVO} were linearly approximated in ORCA \cite{ORCA} and extended to double integrator dynamics in AVO \cite{AVO}. Rufli et al. \cite{cnco} extend RVO to generate $n^{th}$ order continuous trajectories.

Berg et al.~\cite{LQG} and Bareiss et al. \cite{LQR} extend VO to \textit{\color{black}control obstacles} for agents with linear dynamics. Moreover, they demonstrate the algorithm for a large swarm of quadrotors by linearizing the quadrotor dynamics about the hover configuration. This type of linearized model is valid only for small roll and pitch angles about the hover configuration \cite{hoverlimitation} and not for large angular deviation as during high-velocity flights.
Further, control obstacles may also result in non-convex solution space and the new velocity is generally computed from a convex subset of this solution space. Cheng et al.~\cite{MPCORCA} avoid this convex approximation by using an MPC-based method for linear systems. Morgan et al. \cite{Morgan} present a decentralized algorithm based on sequential convex programming (SCP). Due to its higher computational complexity, SCP is not favorable for fast online computation.
\textcolor{black}{Baca et al.~\cite{Baca} present a control and collision avoidance algorithm, where collisions are resolved by priority based altitude variations. In many ways, this approach is complimentary to our method. However, collision avoidance through altitude variations may be unsuitable when large number of agents operate in indoor scenarios due to limits on ceiling height.}
Reactive methods are suitable for dynamic environments because they only use the local position and velocity data for the neighboring agents and obstacles (i.e. state information). However, these reactive methods cannot provide any global guarantees.


\subsection{Downwash}
In dense scenarios, multiple quadrotors may have to maneuver in close proximity to each other. Downwash causes a region of instability below the rotors of a quadrotor and any other quadrotor entering this region may lose control or result in  unstable behavior \cite{Preiss}. In prior literature, the downwash effect is considered in collision avoidance by modeling the agents as axis-aligned ellipsoids \cite{LQR, Preiss} or cylinders \cite{concurrentplanning, downwash2}, which encourages a larger separation along the Z-axis. Quadrotor roll and pitch affect the downwash region, and the ellipsoid or cylinder must be rotated with respect to the quadrotor orientation for accurate modeling \cite{downwash2}. For simplicity, the quadrotors are modeled as axis-aligned ellipsoids and the radius of the ellipsoid/cylinder is increased by a safety threshold to account for roll and pitch \cite{downwash2} \cite{concurrentplanning}. 

\section{Preliminaries}
In this section, we introduce our assumptions, notation, and provide an overview of the concepts of differential flatness and feedforward linearization. We describe the quadrotor model and the non-linear transformation used in our collision avoidance algorithm (Section IV).
\subsection{Symbols and Notation}
The symbols and notations used in this paper are defined in Table \ref{tab:notation}.
\begin{table}[t]
 \caption{\label{tab:notation} Notation and symbols.}
\begin{center}
 \begin{tabular}{|p{1.2cm}|p{6.2cm}|} 
 \hline
 \textbf{Notation} & \textbf{Definition} \\ 
 \hline\hline
 $\mathcal{W}$ & World Frame defined by unit vectors $\mathbf{x_W}$, $\mathbf{y_W}$, and $\mathbf{z_W}$ along the standard X, Y and Z axes \\ 
 \hline
 $\mathcal{B}$ & Body Frame attached to the center of mass, defined by the axes $\mathbf{x_B}$, $\mathbf{y_B}$, and $\mathbf{z_B}$ \\
 \hline
 $\mathbf{r}$ & 3-D position of the center of mass of the quadrotor given by $[x, y, z]$ \\
 \hline
 $\mathbf{v}$ & Velocity of the quadrotor given by $[\dot{x}, \dot{y}, \dot{z}]$\\
 \hline
 $\mathbf{a, j}$ & Acceleration and jerk of the quadrotor given by second and third derivative of position respectively\\
 \hline
 $\phi, \theta, \psi$ & Roll, pitch and yaw of the quadrotor.\\
 \hline
 $\mathbf{R}$ & Rotation matrix of quadrotor body frame ($\mathcal{B}$) w.r.t world frame ($\mathcal{W}$)\\
 \hline
 $\mathbf{T}$ & Net thrust in body fixed coordinate frame\\
 \hline
 $\mathbf{m}$ & Mass of quadrotor\\
 \hline
 ${\boldsymbol{\omega}}$ & Angular velocity in body fixed coordinate frame given by $[p,q,r]$\\
 \hline
 $\mathbf{x}$ & Quadrotor State space\\
 \hline 
 $\mathbf{u}$ & Control input to the quadrotor\\
 \hline
 $\mathbf{u_c}$ & Input to the inner loop controller\\
 \hline
 $\mathbf{VO^\tau_{A|B}}$ & Velocity Obstacle for agent A induced by agent B over a time horizon $\tau$\\
 \hline
 $\mathbf{ORCA^\tau_{A|B}}$ & Collision avoiding velocity set for agent A induced by agent B over a time horizon $\tau$\\
 \hline
 $V_{max}$ & Maximum value of velocity allowed by the quadrotor dynamics\\
 \hline
 $A_{max}$ & Maximum value of acceleration allowed by the quadrotor dyamics\\
 \hline
\end{tabular}
\end{center}
\end{table}
\subsection{Differential Flatness}
A nonlinear system given by ${\bf{\dot{x} = f(x,u)}}$ is differentially flat if there exists a set ${\boldsymbol{\zeta}}$ (flat output) whose elements, expressed as $\boldsymbol{\zeta} = [\zeta_1, \zeta_2, ..., \zeta_m]$, are differentially independent; $\boldsymbol{\zeta}$ and their derivatives can be used to construct the system state space and control inputs  \cite{Fliess95} \cite{ExactFeedForward}. 
The quadrotor model we consider is described below and, from \cite{Mellinger}, we know that the quadrotor dynamics is differentially flat for the flat output set given by  $\boldsymbol{\zeta} = [\zeta_1, \zeta_2, \zeta_3, \zeta_4] = [x,y,z,\psi]$.\\

\noindent{\bf{Quadrotor Model:}}
The state space and the control input for the quadrotor are given by
\begin{eqnarray}
{\bf{x}} = [x, y, z, \dot{x}, \dot{y}, \dot{z}, \phi, \theta, \psi, p, q, r],\\
{\bf{u}} = [T, \dot{\phi}, \dot{\theta}, \dot{\psi}].
\end{eqnarray}
The quadrotor dynamics can be represented by the following set of equations:
\begin{eqnarray}
\label{eqn:rfot=v}
    \mathbf{\dot{r}} = \mathbf{v},\\
\label{eqn:accel}
    \mathbf{m\bar{a} = -mgz_{W} + Tz_{B}},\\
\label{eqn:rdot}
    \mathbf{\dot{R} = R \times \boldsymbol{\omega}^T},\\
\label{eqn:tau}
    \boldsymbol{\dot{\omega}} = \mathbf{J^{-1}}[-\boldsymbol{\omega} \times \mathbf{J}\boldsymbol{\omega} + \boldsymbol{\tau}].
\end{eqnarray}

\subsection{Feedforward Linearization}
Hagenmeyer et al.~\cite{ExactFeedForward} introduce the notion of exact feedforward linearization based on differential flatness. Given a non-linear, differentially flat system ${\bf{\dot{x} = f(x,u)}}$ and a sufficiently smooth trajectory in the flat output $\zeta_{ref}$, the system can be represented as a linear flat model given as 
\begin{align}
    \label{eqn:flatstate}
    {\boldsymbol{\dot{\xi}}} = \boldsymbol{A{\xi}} + \boldsymbol{B\nu},\\
    \label{eqn:nu}
    \boldsymbol{\nu} = \psi(\boldsymbol{\xi}, \mathbf{u}, \mathbf{\dot{u}}, ..., \mathbf{u^\sigma}),
\end{align}
when a nominal input (\ref{nom_input}) is applied to the non-linear system, provided the initial condition (\ref{eqn:InitialCondition}) is satisfied.
\begin{eqnarray}\label{nom_input}
    \mathbf{u^*} = \psi^{-1}(\boldsymbol{\xi}_{desired}, \boldsymbol{\nu}_{desired}),\\
\label{eqn:InitialCondition}
    \boldsymbol{\xi}(0) = \boldsymbol{\xi}_{desired}(0).
\end{eqnarray}
Here, $\xi$ is the linear multi-variable Brunovsky form, represented as
\begin{align}
   \boldsymbol{\xi} = [\zeta_1, \dot{\zeta_1}, ..., \zeta_1^{\rho_1 -1}, \zeta_2, ..., \zeta_{m-1}, \dot{\zeta_{m-1}}, ..., \zeta_{m-1}^{\rho_m -1}],
\end{align}
and $\nu$ represents the new control input in the flat space.
The desired flat state $(\boldsymbol{\xi}_{desired}$) and flat input ($\boldsymbol{\nu}_{desired}$) can be computed from $\zeta_{ref}$. In Equation (\ref{eqn:nu}), $\sigma$ is the maximum order of differentiation of $\mathbf{u}$ required to describe $\boldsymbol{\nu}$.

Equation (\ref{nom_input}) represents the non-linear transformation required to obtain $\bf{u}$ from the flat input. We observe from {\color{black}Etal et al. \cite{Etal}} that we require the third derivative of position ({\bf{r}}) and the first derivative of yaw ($\psi$) to express the state and the control input. Hence, we choose the flat state space $\mathbf{z}$ and the flat input $\boldsymbol{\nu}$ as
\begin{align*}
    \mathbf{z} = [\mathbf{r}, \mathbf{v}, \mathbf{a}, \psi],\\
    \boldsymbol{\nu} = [\mathbf{j}, \dot{\psi}].
\end{align*}
In our method, flatness-based feedforward linearization provides the benefit of reducing the computation overload by linearizing the quadrotor dynamics to linear flat model while still allowing us to handle the non-linearities using a non-linear map. 
We use a flatness-based MPC that is similar to the one used in \cite{Greeff} to generate a feasible collision-free trajectory. The non-linear map (Eqn. \ref{eqn:non-linear_map}) is described below:

{\noindent\bf{Non-linear Map: }}
The control input {\bf{u}} can be represented in terms of $\zeta$ and its derivative using the following relation,
\begin{equation}\label{eqn:thrust}
    u_1 = T = m\dfrac{g+\ddot{\zeta_3}}{\mathbf{R_{33}}},
\end{equation}
\begin{equation}
\begin{bmatrix}
u_2\\
u_3\\
u_4
\end{bmatrix}
=
\begin{bmatrix}
\dot{\phi}\\
\dot{\theta}\\
\dot{\psi}
\end{bmatrix}
= \frac{1}{T}
\begin{bmatrix}
    -Y_{B}^T\\
    {-X_{B}^T}/{cos{\phi}}\\
    0
\end{bmatrix}
j + \dot{\psi}_{ref}
\begin{bmatrix}
sin{\theta}\\
{-cos{\theta}}{tan{\phi}}\\
1
\end{bmatrix}.
\label{eqn:non-linear_map}
\end{equation}
Eqn. \ref{eqn:non-linear_map} gives the non-linear map that transforms the flat inputs to the quadrotor inputs \cite{Etal}.

We assume the presence of an inner-loop attitude controller that can track the attitude values and takes as input ${\bf{u}}_{c} = [T, {\phi_{cmd}}, {\theta_{cmd}}, \dot{\psi}]$. The inner loop control dynamics is given by
\begin{eqnarray}\label{eqn:innerloop1}
    \dot{\phi} = \frac{1}{\tau_{\phi}}(K_{\phi}\phi_{cmd}-\phi),\\
\label{eqn:innerloop2}
    \dot{\theta} = \frac{1}{\tau_{\theta}}(K_{\theta}\theta_{cmd}-\theta),\\
\label{eqn:innerloop3}
    \dot{\psi} = \psi_{ref}.
\end{eqnarray}
From Eqn. (\ref{eqn:thrust}) - (\ref{eqn:innerloop3}) we can compute ${\bf{u}_{c}}$ in terms of ${\boldsymbol{\zeta}}$ and its derivative, which is then applied as input to the inner loop controller.

\subsection{Assumptions on Swarm Agents}
We assume that each agent has access to a reference trajectory in flat output space $\boldsymbol{\zeta_{ref}}=[x,y,z,\psi]$ that considers static obstacles in the environment. The reference trajectory is assumed to be sufficiently smooth and can be generated prior to flight using any trajectory generation method, such as \cite{Liusearch}. In our case, we assume the yaw orientation of the quadrotor is not important and we take $\psi = 0$ in the reference trajectory. Though the reference trajectory considers static obstacles, collision avoidance can deviate the agent from its reference trajectory causing them to collide with static obstacle. Such issues are managed by constructing VO with the closest point obstacle, as shown in \cite{LSwarm}. We assume the availability of position, velocity, and orientation of each neighboring quadrotor/obstacle to each swarm agent at any given time for generating ORCA constraints. Our collision avoidance scheme is based on ORCA~\cite{ORCA} and \cite{MPCORCA}, and we assume that neighboring agents and dynamic obstacle travel at a constant velocity during the prediction horizon of the MPC.
\section{Collision Avoidance and Trajectory Computation}
In this section, we present our decentralized collision avoidance algorithm for the quadrotor swarm. 
In our algorithm, the quadrotor dynamics are handled in the flat-MPC, while ORCA planes are used as state constraints to generate local, collision-free, downwash-aware trajectories. The high-level overview of our algorithm is given in Fig. \ref{block_diagram} and the details are given below.


\subsection{MPC}
Model Predictive Control (MPC) is a receding horizon planner that computes a control input based on the system dynamics, input and state constraint, by minimizing an objective function over a prediction horizon. Here, prediction horizon (N) is a finite time horizon in the future [t, t+N]. In our method, we linearize the quadrotor model using feedforward linearization, as mentioned in Section III, to facilitate the use of a linear MPC.
The linearized flat state space $\mathbf{z}$ and the flat input $\boldsymbol{\nu}$ are given by
\begin{eqnarray*}
\mathbf{z} = [\mathbf{r}, \mathbf{v}, \mathbf{a}, \psi],\\    
\boldsymbol{\nu} = [\mathbf{j}, \dot{\psi}].
\end{eqnarray*}


At each time step, the MPC uses the linear flat model ({\bf{z}}) to plan the state trajectories in the flat space and to compute a control input in the flat input space $\boldsymbol{\nu}$. The optimization problem minimizes the tracking error and flat input ($\boldsymbol{\nu}$) while satisfying the velocity constraints by ORCA and constraints on jerk. 
The optimization problem is given by
\begin{equation}
\begin{aligned}
& \underset{}{\text{minimize}} \quad
& & \sum_{t=0}^{N} \left(\boldsymbol{\zeta_{ref,t}} - \boldsymbol{\zeta_{t}}\right)Q\left(\zeta_{ref,t} - \zeta_{t}\right) + \mathbf{\nu}_{t}R\mathbf{\nu}_{t} \\
& \textrm{subject to} \quad
& & \xi_0 = \xi_t\\
& & & \xi_{k+1} = A\xi_{k} + B\nu_{k}\\
& & & |v_{x,k}| \le V_{max,x}, |v_{y,k}| \le V_{max,y}, |v_{z,k}| \le V_{max,z}\\
& & & |a_{x,k}| \le A_{max,x}, |a_{y,k}| \le A_{max,y}, |a_{z,k}| \le A_{max,z}\\
& & & \text{ORCA}\left(C\xi_{k+1}, {C\xi_{neighbour, k+1}}\right ) \le 0\\
& & & {J\left(\xi_k, \phi_{min}, \phi_{max}, \theta_{min}, \theta_{max}\right)} \le 0\\
& & & \forall k=0, 1, ..., N-1.
\end{aligned}
\label{eqn:optimization}
\end{equation}

Here `$N$' is the number of prediction steps, $\zeta_{ref}$ is the reference trajectory, and the matrices Q and R are weights that prioritize between minimizing the trajectory tracking error and the control input. $\xi_k$ represents the flat state of the agents and matrix $C$ is such that $C\xi_{k}$ gives the position, velocity and orientation of the agent at time step k. $\xi_{neighbour,k}$ is the flat state of the neighboring agent/obstacle at time step k. 
During collision avoidance, the quadrotor may have to deviate from its reference trajectory to avoid collision. In this case, the initial state of the quadrotor agent may deviate from the reference trajectory, the state feedback is used to satisfy the initial condition requirement given by ({\color{black}\ref{eqn:InitialCondition}}). Equation $\xi_0 = \xi_t$ represents the state feedback. The ORCA velocity constraints between the agent and its neighbors are represented by $ORCA(C\xi_{k+1}, {C\xi_{neighbour, k+1}}) \le 0$. The constraints on the jerk are represented by ${J(\xi_k, \phi_{min}, \phi_{max}, \theta_{min}, \theta_{max})} \le 0$, and are computed as follows.

\begin{figure}[t]
\centering
\includegraphics[height=1.5in, width=3.0in]{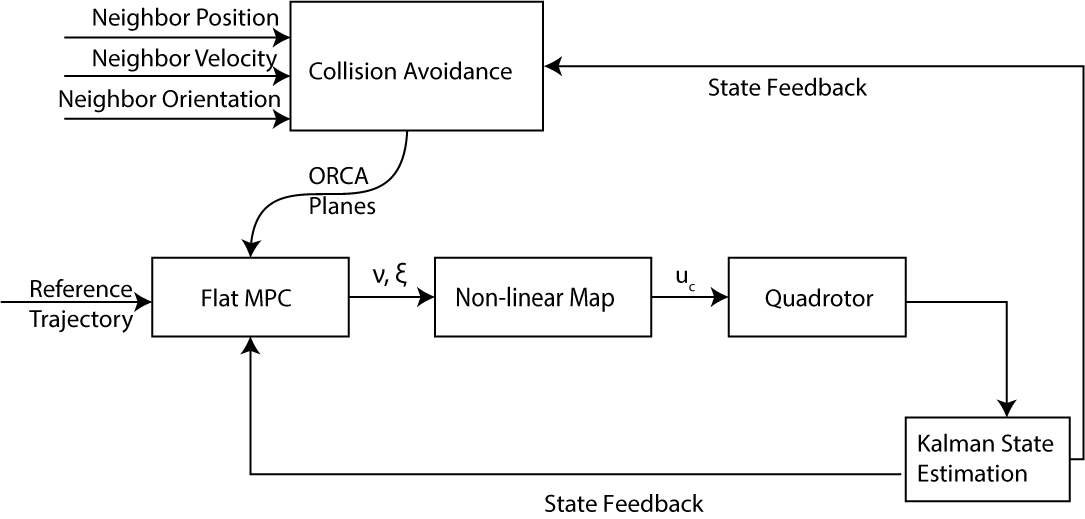}
\caption{\color{black} Collision avoidance framework for each agent: The Flat-MPC incorporates the ORCA velocity constraints as state constraints in the the optimization problem and computes a feasible trajectory for the agents. The position, velocity, and orientation of an agent and neighbours are used in the ORCA block to compute a collision avoidance velocity constraint that accounts for downwash.}
\label{block_diagram}
\end{figure}
\noindent \textbf{Constraints on Jerk:}
Assume the quadrotor can reach a maximum and minimum roll and pitch angle given by ($\phi_{max}, \phi_{min}$), and ($\theta_{max}, \theta_{min}$). Since the MPC computes a control input in flat input $\nu$, the constraint on roll and pitch need to be transformed to the flat space to prevent quadrotors from flipping over during collision avoidance. At any timestep, the maximum jerk that can be applied to the system is dependent on the current state of the quadrotor, including maximum and minimum limits on roll and pitch.

A maximum and minimum value for the roll rate ($\dot{\phi}_{max}, \dot{\phi}_{min}$) and pitch rate ($\dot{\theta}_{max}, \dot{\theta}_{min}$) can be computed using $\phi_{max}, \phi_{min}, \theta_{max}, \theta_{min}, \xi$, Eqn. (\ref{eqn:innerloop1}) and (\ref{eqn:innerloop2}).
To compute the limits on jerk ($j_{min}, j_{max}$) at a given timestep, 
we re-write Eqn. (\ref{eqn:non-linear_map}) as follows. The minimum jerk can be computed as
\begin{equation*}
j_{min} 
=
\min_{\dot{\phi}, \dot{\theta}}\hspace{0.2cm}
T
\begin{bmatrix}
    -Y_{B}^T\\
    {-X_{B}^T}/{cos{\phi}}
\end{bmatrix}^{-1}
\left (
\begin{bmatrix}
\dot{\phi}\\
\dot{\theta}
\end{bmatrix} 
-
\dot{\psi}_{ref}
\begin{bmatrix}
sin{\theta}\\
{-cos{\theta}}{tan{\phi}}
\end{bmatrix}
\right ).
\end{equation*}
Similarly, we can compute $j_{max}$ by maximizing the above expression. The computed $j_{max}$ and $j_{min}$ values are used as maximum and minimum jerk constraints, respectively, over the entire prediction horizon.

\noindent\textbf{MPC Output:} As a result of the optimization, the MPC generates the predicted state trajectory and computed control input for each time step in the time horizon. The values for the state and the control input ($\nu$) are passed to the non-linear map to compute the control input for the inner loop attitude controller ($u_{c}$) using the non-linear map (\ref{eqn:non-linear_map}). 

\subsection{\color{black}Collision Avoidance}
ORCA generates linear collision avoidance constraints, which result in a convex set of feasible velocity \cite{ORCA}. In our method, we incorporate ORCA velocity constraints as state constraints in our MPC formulation. 

\subsubsection{State Constraints}\label{stateconstraint}
At a time t = 0, ORCA constraints can be computed for an agent considering the current position and current velocity of all its neighboring agents and obstacles. For subsequent timesteps $\{t+i \mid i = 1,2...N-1\}$, we compute ORCA planes by predicting the future position and velocity of the agents and neighbours using their current state information. 
Assume that we are to compute the ORCA constraints for an agent A.
The position and velocity of agent A is obtained from the state trajectory predicted by the MPC. This is given by $$\{\xi_{predicted,k}\mid k = 1,2,...N\}.$$
In contrast, for neighboring agents we propagate their current position using a simple motion model given by 
$$\mathbf{r_{t+1}} = \mathbf{r_{t}} + \mathbf{v_{t}}t.$$
We assume velocity for the neighboring agents and obstacles to remain constant over the prediction horizon. 
State constraints by ORCA now pertain to a particular time step in the prediction horizon. The ORCA planes consider downwash and uncertainty in the sensor reading. The modifications to ORCA for downwash is presented below. 

\subsubsection{Downwash}
In dense scenarios, multiple quadrotors may have to maneuver in close proximity to each other. Downwash causes a region of instability below the rotors of a quadrotor and any other quadrotor entering this region may lose control and face instability. To ensure safe flights in close proximity, we must account for downwash in our collision avoidance method.



Since we do not make small angle assumptions regarding the attitude angle, assuming a fixed orientation in the form of axis-aligned ellipsoids is not ideal. Hence, we rotate the ellipsoidal bounding region of the agent according to the quadrotor attitude when computing the VO. Two issues require proper consideration in this approach. 
\begin{enumerate}
    \item When two quadrotors are in proximity, only the quadrotor (and its orientation) at the higher altitude influences the downwash region. The quadrotor at the lower altitude, in spite of its orientation, may face instability when it enters the downwash region of the quadrotor above it. 
    \item ORCA requires Minkowski sum construction and closest point computation for constructing the ORCA half-planes. Modelling the quadrotors as oriented ellipsoids increases the complexity of both the computation. Since we are required to recompute the ORCA plane over the prediction horizon as discussed in Section \ref{stateconstraint}, we need a fast method to perform this computation.
\end{enumerate}

We solve the issue by modelling only one of the two quadrotors as ellipsoids while constructing the VO for the pair of agents. To maintain the symmetry of $VO^\tau_{A|B}$ and $VO^\tau_{B|A}$ about the origin, the decision regarding which quadrotor is modelled as an ellipsoid must be common for any pair of quadrotors.

When an agent constructs the VO based on its neighbors, we choose to model the quadrotor at the higher altitude (at the current timestep) as an ellipsoid, while the other is modelled as a sphere. Also, the orientation of the agent at the higher altitude is used to rotate the bounding ellipsoid for the construction of the Minkowski sum. This results in the Minkowski sum being computed between a oriented ellipsoid and a sphere. Given two agent 'i' and 'j', assume agent 'j' is at a higher altitude at the current timestep. Then the Minkowski sum is given by 
$$ {\cal O}_{ij} = {\cal O}_{i,sphere} \oplus {\cal O}_{j,ellipsoid, \phi, \theta}.$$

In addition, since the choice of agent at a higher altitude is unique for any pair of agents, this ensures that $VO^\tau_{A|B}$ and $VO^\tau_{B|A}$ are symmetric about the origin.

\subsection{Modeling Uncertainty}
\begin{figure}[t]
      \centering
      \includegraphics[height=2.0in, width=3.25in]{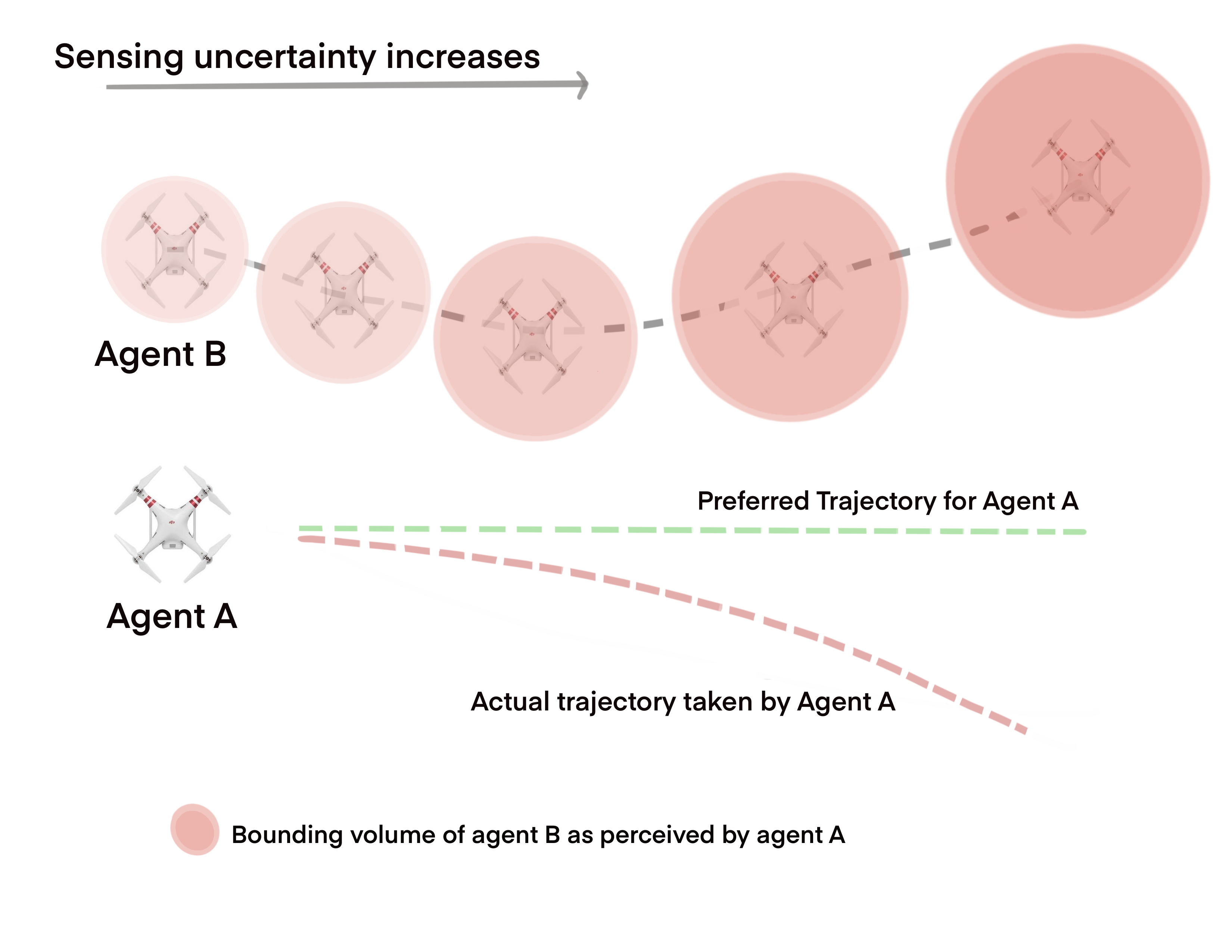}
      \caption{Top view of two quadrotors A and B flying close to each other is shown. From left to right, the uncertainty in the position and velocity of B as sensed by A increases. To account for the uncertainty, A increases its perceived bounding ellipsoid of agent B and computes its new, dynamically feasible velocities based on B's increased size. Our method makes A take a conservative trajectory (red) away from B, rather than its preferred trajectory (green).}
      \label{uncertainty}
\end{figure}
ORCA assumes perfect knowledge of the sizes, positions and velocities of all the obstacles around them. This assumption is idealistic and makes implementing ORCA on real quadrotors impractical. To account for a quadrotor's imperfect sensing, we use a Kalman filter to compute the mean and covariance of the position and velocity of all obstacles around it. \\
The agent's process and observation model is assumed to be as follows,
\begin{align*}
    \mathbf{x_{k+1} = Ax_{k} + Bu_{k} + w_k}\\
    \mathbf{y_k = Cx_k + v_k}
\end{align*}
Here $\mathbf{x_k}$ is the state and $\mathbf{u_k}$ is the control input at time step k. $\mathbf{w_k}$ and $\mathbf{vk}$ are the process and measurement noise respectively and are assumed to be Gaussian, zero-mean white-noise. The state transition matrix $\mathbf{A}$, matrix $\mathbf{B}$ and $\mathbf{C}$ are same as in the linear flat model we compute for the quadrotor. The mean and the co-variance for the states are computed through kalman predict and update cycle. 

Eigenvalues of the covariance matrix represent the spread of the data, or the level of uncertainty in the direction of the eigenvectors. Therefore, we use the maximum eigenvalue of an entity's position covariance matrix to enlarge its bounding volume from the perspective of the quadrotor that senses it. If the sensed entity is another agent, we increase the length of the axes of its bounding ellipsoid and, if the entity is a dynamic obstacle, we increase the radius of its bounding sphere. {\color{black}This is similar to the method used in \cite{Snape}. The VO is constructed using this bounding sphere and is later augmented by the velocity covariance matrix}. Although this formulation makes the collision avoidance conservative, we observe that it works well even with tens of dynamic obstacles. Fig. \ref{uncertainty} shows a scenario with two agents and the increase in the size of an agent's bounding ellipsoid as perceived by another agent. This is a simple approach to account for uncertainty, which could be extended to more sophisticated methods, as shown in \cite{Gopalakrishnan} or \cite{Kamel}.

\section{Results}
In this section we highlight our implementation and the experimental results and describe the benefits of our method over prior methods.
\subsection{Experimental Setup}
Our algorithm is implemented on an Intel Xeon w-2123 octacore processor ($3.6$ GHz) with $32$ GB memory and GeForce GTX $1080$ GPU. We utilize the PX4 Software In The Loop framework, ROS Kinetic, and Gazebo $7.14.0$ for our simulations. The RVO-3D library is utilized to compute the ORCA collision avoidance constraints. 
The Optimal Control Problem (OCP) is solved using IPOPT Library with a prediction horizon of 10 steps and a timestep of 0.1 seconds. 
We consider a sensing region of 6m and a time horizon of 5 seconds for ORCA (including the one used in our formulation), AVO, and LQR-obstacle. The $\delta$ parameter for AVO is set as $2v_{max}/a_{max}$ as suggested in \cite{AVO}.

\subsection{Performance Evaluation}
We compare the performance of our algorithm with ORCA, AVO and LQR-Obstacles in terms of smoothness of trajectories, variation in velocity during collision avoidance, and proportion of collisions while maneuvering trajectories with high-velocity.
In addition, we show the separation between agents to demonstrate the downwash performance.

\subsubsection{Generated Trajectory}
\begin{figure*}[t]
   \centering
   \begin{subfigure}[b]{0.32\textwidth}
   \includegraphics[width=1\linewidth]{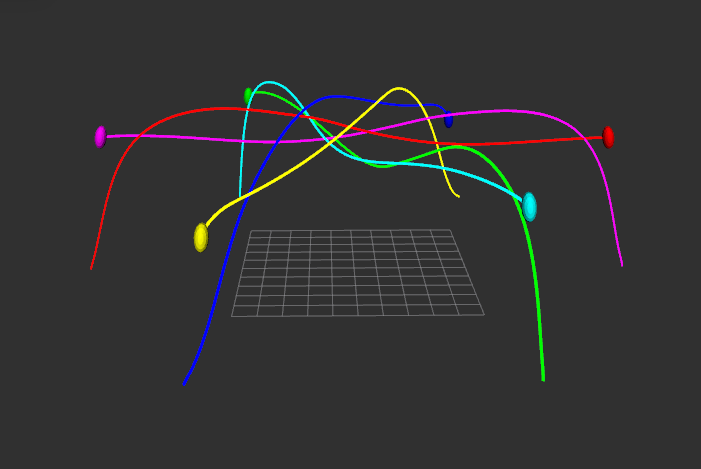}
   \caption{Proposed Method}
   \label{fig:New_traj}
   \end{subfigure}
   \begin{subfigure}[b]{0.32\textwidth}
   \includegraphics[width=1\linewidth]{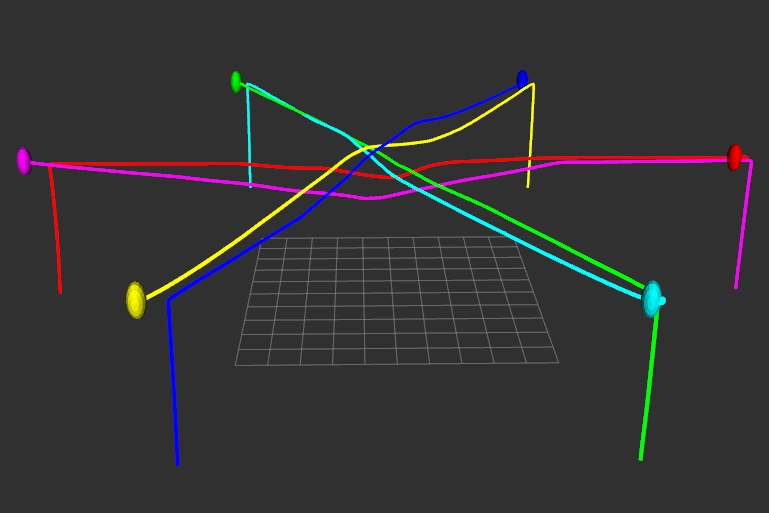}
   \caption{ORCA}
   \label{fig:Orca_traj}
   \end{subfigure}
   \begin{subfigure}[b]{0.32\textwidth}
   \includegraphics[width=1.0\linewidth]{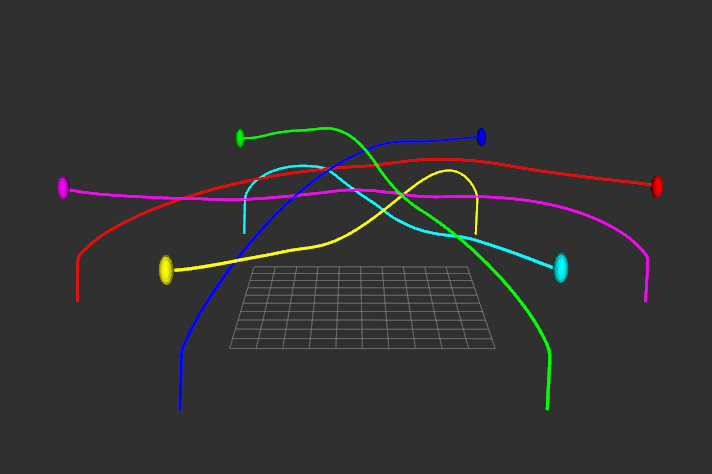}
   \caption{AVO}
   \label{fig:Avo_traj}
   \end{subfigure}
   \caption{The trajectories generated using a) our proposed method b) ORCA and c) AVO, when 8 agents exchange their position with diagonally adjacent agents. The colored circles represent the starting position of the agents. Each colored trajectory represents a different agent. The quadrotors travel at a max velocity of 5m/s}
   \label{fig:traj}
\end{figure*}
We compare the smoothness of the trajectories followed by 6 agents while using our method, ORCA, and AVO. The agents are initially in a circular formation, exchanging their positions with diagonally opposite agents. The maximum velocity for the agents in the above scenario was limited to 4 m/s. Figure \ref{fig:New_traj}, \ref{fig:Orca_traj}, and \ref{fig:Avo_traj} illustrate the trajectories followed by the 6 agents using our method, ORCA and AVO.  We observe that our method produces smoother trajectories than ORCA.

\subsubsection{Variation in Velocity}
\begin{figure*}[t]
   \centering
   \begin{subfigure}[b]{0.32\textwidth}
   \includegraphics[width=1\linewidth]{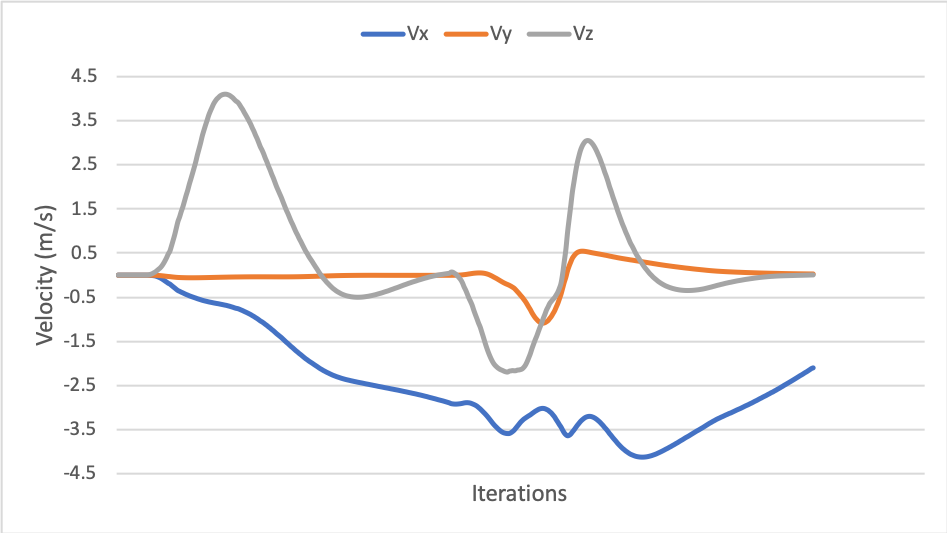}
   \caption{Proposed Method}
   \label{fig:proposedMethod_Vel}
   \end{subfigure}
   \begin{subfigure}[b]{0.32\textwidth}
   \includegraphics[width=1\linewidth]{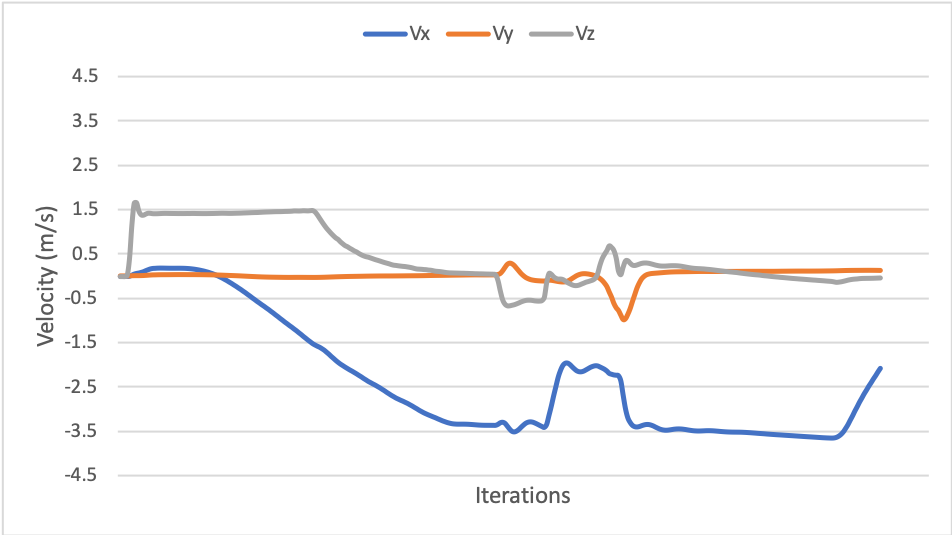}
   \caption{ORCA}
   \label{fig:ORCA_Vel}
   \end{subfigure}
   \begin{subfigure}[b]{0.32\textwidth}
   \includegraphics[width=1\linewidth]{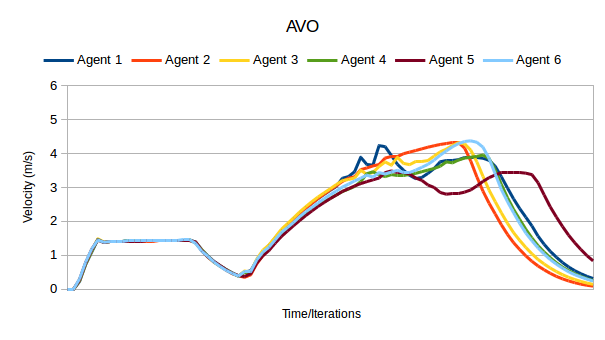}
   \caption{AVO}
   \label{fig:AVO_Vel}
   \end{subfigure}
      \caption{Variation in velocity when agents exchange their positions to antipodal positions in circular scenarios when collision avoidance is performed using a) proposed method, b) ORCA, and c) AVO. Our scenarios consist of 6 quadrotors with a maximum velocity of 5m/s. }
      \label{fig:velvar}
\end{figure*}
To show our method results in smoother trajectories than AVO, we plot the velocity components (along X, Y, and Z axes) of an agent while performing collision avoidance in the scenarios discussed in the previous subsection. Figure \ref{fig:velvar} graphs the variation in velocity of one agent while using our method, ORCA, and AVO. It can be observed that the variation velocity is much smoother for the agent using our method compared to ORCA or AVO.

\subsubsection{Performance During Agile Maneuvers}
Table \ref{tab:highacc} summarizes the performance of the various algorithms while following high velocity time-parameterized trajectories. The trajectories provided are straight lines, but the quadrotors accelerate to reach their highest velocity in the midpoint of the line before decelerating. We define an episode as when all 8 quadrotors exchange their positions with the agents diagonally opposite them. We report the number of episodes (out of 250 episodes) that resulted in one or more of the agents colliding. We observe that our method performs better when following such trajectories.\\

\begin{table}[t]
  \centering
  \begin{tabular}{|p{1cm}||p{1.5cm}|p{1.5cm}|p{3.0cm}|}
   \hline
 \multicolumn{4}{|c|}{Collisions while tracking a reference trajectory} \\
 \hline
 Average Velocity & \multicolumn{3}{|c|}{Episodes with Observed Collisions (out of 250 Episodes)}\\
 \cline{2-4}
 &  ORCA & AVO & DCAD (Our Method)\\
 \hline
 \hline
 1 m/s  &  0 & 0 & 0\\
 \hline
 2 m/s  &  0 & 0 & 0\\
 \hline
 4 m/s  & 17 &31 & 0\\
 \hline
 7 m/s  &151 &128 & 21\\
  \hline
  \end{tabular}
  \caption{7m  Number of episodes in which one or more quadrotor collided when the agents used ORCA, AVO, LQR or our method for collision avoidance. The scenarios consisted of 8 quadrotor in a circle exchanging their positions with the antipodal agents. It can be observed that in high-velocity trajectories, our method shows good performance.}
  \label{tab:highacc}
\end{table}

In previous literature, ORCA and AVO are generally tested in scenarios where only the goal position is provide, this is in contrast to using a time parameterized trajectory as in our previous result. Hence, we also tabulate the results when only a goal position is provided in Table \ref{tab:highvelgoal}. The scenarios is same as earlier, the agents exchange positions with the diagonally opposite neighbour.

\begin{table}[t]
  \centering
  \begin{tabular}{|p{1cm}||p{1.5cm}|p{1.5cm}|p{2.5cm}|}
  \hline
 \multicolumn{4}{|c|}{Collisions while moving to a goal position} \\
 \hline
 Average Velocity & \multicolumn{3}{|c|}{Episodes with Observed Collisions (out of 250 Episodes)}\\
 \cline{2-4}
 &  ORCA & AVO & DCAD (Our Method)\\
 \hline
 \hline
 1 m/s  &0 & 0 & 0\\
 \hline
 2 m/s  &0 & 0 & 0\\
 \hline
 4 m/s  &0  &21 & 0\\
 \hline
 7 m/s  &67 &106 & 21\\
  \hline
  \end{tabular}
  \caption{Number of episodes in which one or more quadrotor collided when the agents used ORCA, AVO, LQR or our method for collision avoidance. The scenarios consisted of 8 quadrotor in a circle exchanging their positions with the antipodal agents. It can be observed that in high velocity our method shows good performance.}
  \label{tab:highvelgoal}
\end{table}

Comparing with both cases, we can observe that our method performs better in high velocity flight in scenarios with small sensing distance (6m in our case). 

\begin{figure}[thpb]
   \centering
   \begin{subfigure}[b]{0.47\textwidth}
   \centering
   \includegraphics[width=0.8\linewidth, height=1.5in]{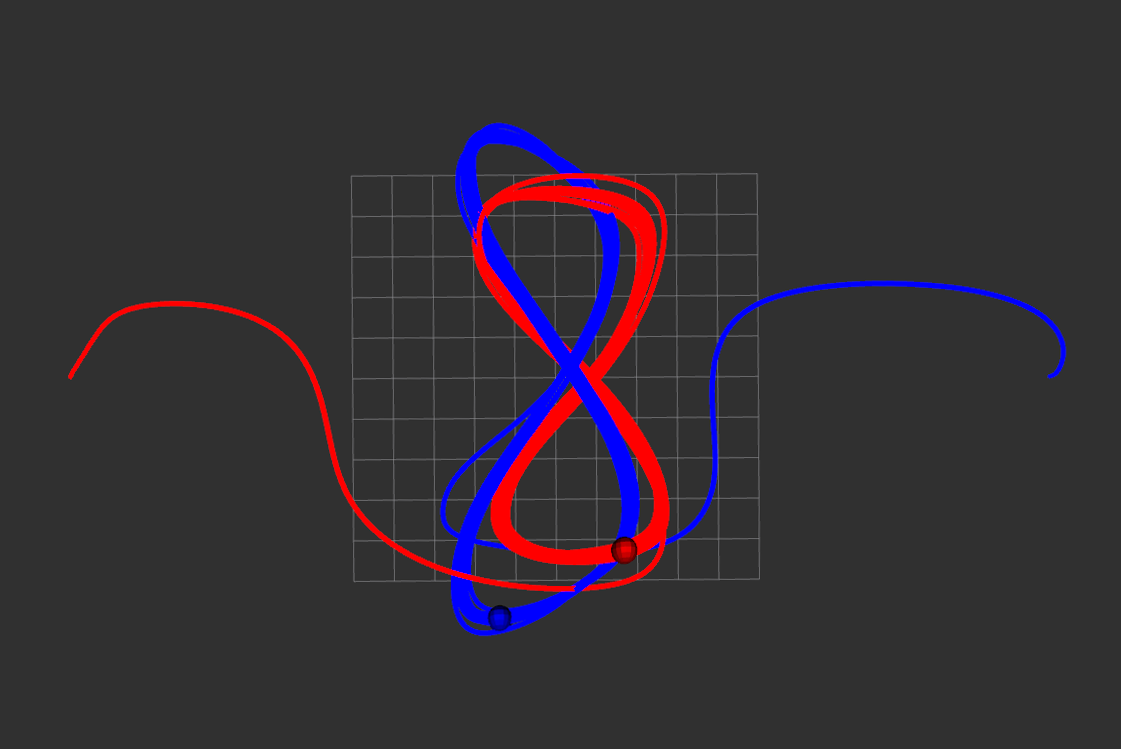}
   \caption{Top View}
   \label{fig:lamniscate_nm_top}
   \end{subfigure}
   \begin{subfigure}[b]{0.47\textwidth}
   \centering
   \includegraphics[width=0.8\linewidth, height=1.5in]{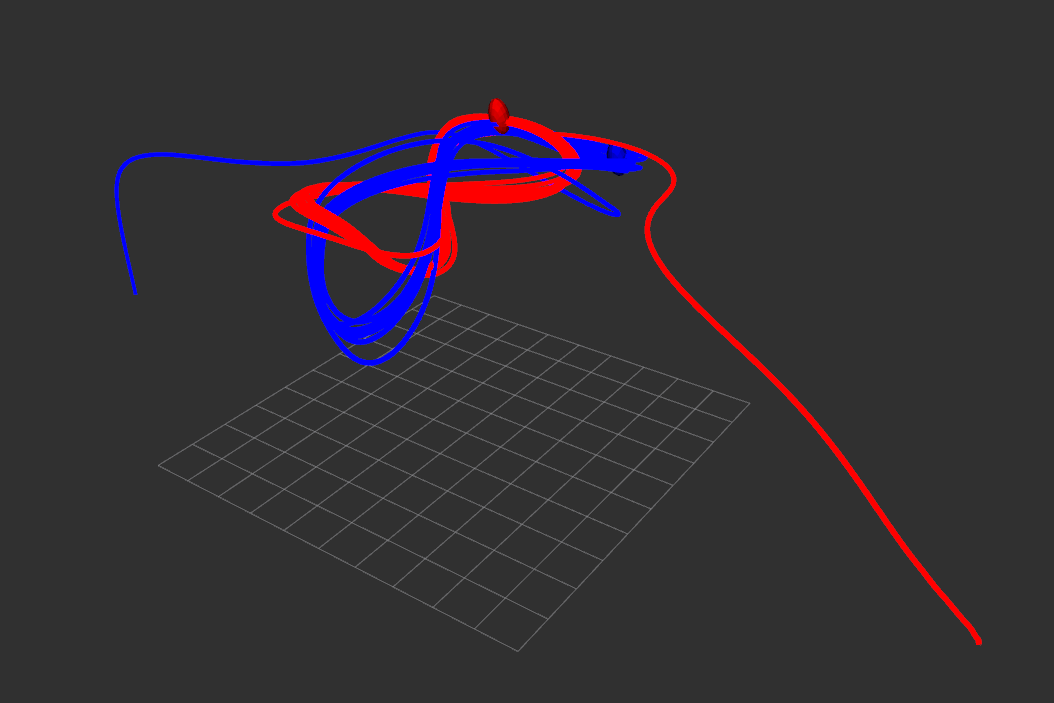}
   \caption{Bird's Eye View}
   \label{fig:lamniscate_nm_be}
   \end{subfigure}
   \caption{Two quadrotors (represented by red and blue) tracking the same lamniscate trajectory in opposing directions (red moving clockwise and blue moving anti-clockwise) at an average speed of 4.2 m/s. We observe that quadrotors alter their trajectories to avoid the collisions even while performing such aggressive maneuvers.}
   \label{fig:laminscate}
\end{figure}

In addition, we observe that our collision avoidance performs well in aggressive trajectories such as a lamniscate (shown in figure \ref{fig:laminscate}). 

\subsubsection{Optimality}
\textcolor{black}{In order to compare the optimality of the generated trajectories we consider two metrics, namely, length of trajectory generated and time to reach goal. We consider a circular scenario as before and the tabulate the average trajectory length and the mean time taken to reach the goal. The agents move to their anti-podal positions which are 40m away. Hence, in the absence of any obstacle the optimal path would be a straight line path with a distance of 40m to the goal. From table \ref{tab:optimality} we can observe that the trajectory length while performing collision avoidance using DCAD is close to 40m as in the ideal obstacle-free case. Similarly the path lengths are comparable to other methods such as AVO, ORCA. Further, it can be noted that the time to reach the goal is approximately the same.}
\begin{table}[t]
  \centering
  \resizebox{\columnwidth}{!}{%

  \begin{tabular}{|c|c|c|c|c|c|c|}
    \hline
   \multirow{2}{*}{Number of Agents} & \multicolumn{3}{c|}{Trajectory Length (m)} & \multicolumn{3}{c|}{Time to Goal (s)}\\
   \cline{2-7}
   & ORCA & AVO & DCAD & ORCA & AVO & DCAD\\
   \hline
   2 & 40.62 & 40.75 & 41.08 & 15.71 & 15.71 & 15.71\\
   4 & 40.77 & 40.96 & 41.42 & 15.71 & 15.71 & 15.70\\
   6 & 40.84 & 41.74 & 41.53 & 15.72 & 15.71 & 15.71\\
   8 & 41.55 & 41.94 & 42.34 & 15.70 & 15.72 & 15.71\\
   10 & 41.96 &42.17 & 42.97 & 15.70 & 15.71 & 15.70\\
   \hline

  \end{tabular}
  }
  \caption{Comparison of mean path length and mean time taken to reach goal when the agents use ORCA, AVO, and DCAD. A circular scenario is considered when agents exchange positions with their diagoannly opposite agents. The circle is of diameter 40m and therefore the optimal path length in an obstacle-free case would be 40m.}
  \label{tab:optimality}
\end{table}

\subsubsection{Scalability Comparison}
\begin{figure}[t]
   \centering
   \includegraphics[height=2in, width=0.47\textwidth]{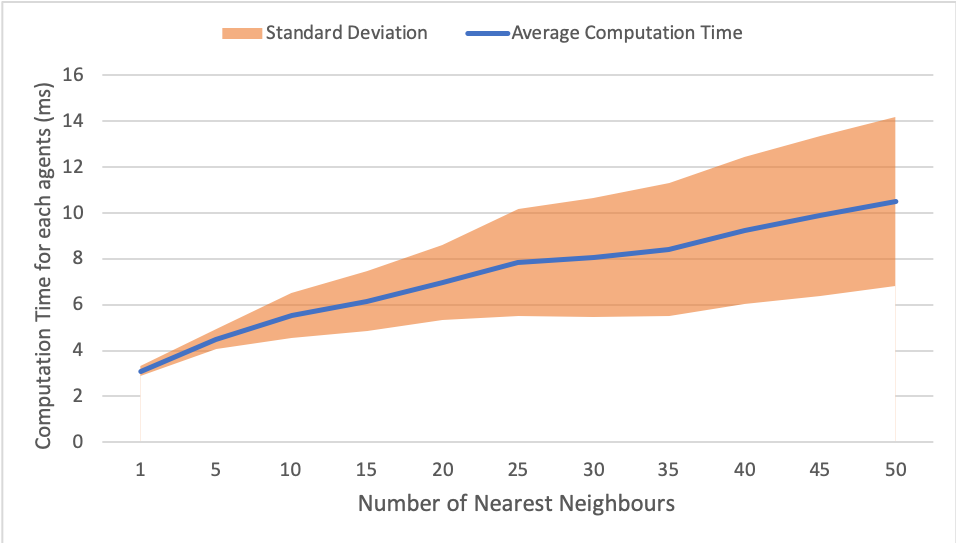}
   \caption{The time (ms) required to compute a collision-free trajectory for a single agent per timestep considering 1 to 50 nearest neighbors. }
   \label{fig:scalability}
\end{figure}

The Figure \ref{fig:scalability} illustrates the computation time of our algorithm for one agent considering 1 to 50 obstacles in the environment. We observe that considering only the closest 10 obstacles provides good performance in most cases, an on an average this take less than 6ms.

\begin{figure}[tbhp]
   \centering
   \includegraphics[height=2in, width=0.47\textwidth]{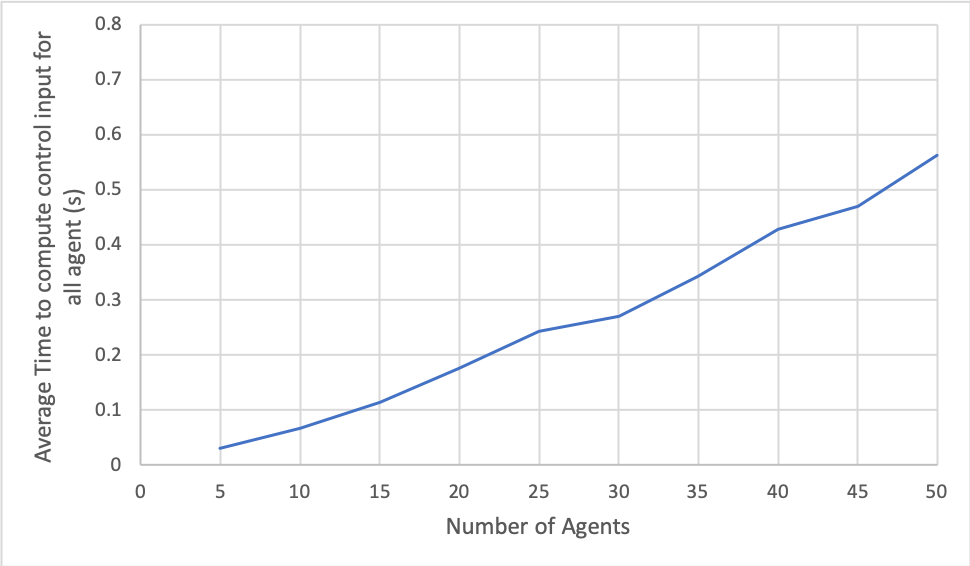}
   \caption{The time (ms) required to compute a collision-free trajectory for a single agent per timestep considering 1 to 50 nearest neighbors. }
   \label{fig:scal_all}
\end{figure}

Figure \ref{fig:scal_all} illustrates the average time taken by our algorithm for computing control input for all agents in the scenario presented in Fig. \ref{fig:traj}. The number of agents in the scenario is varied from 5 to 50 agents. Each agent considers only the nearest 10 agents as obstacles.

Table \ref{tab:compTime} below tabulates the average time taken by DCAD and ORCA for computing control input. The number of agents in the scenario is varied from 5 to 50 agents. Each agent considers the nearest 10 agents as obstacles.

\begin{table}[t]
  \centering

  \begin{tabular}{|c|c|c|}
    \hline
   \multirow{2}{*}{Number of Agents} & \multicolumn{2}{c|}{Computation Time (ms)} \\
   \cline{2-3}
   & ORCA & DCAD\\
   \hline
    1 & 0.009 & 5.1\\
    5 & 0.023 & 29.8\\
   10 & 0.097 & 66.6\\
   15 & 0.141 & 113.3\\
   20 & 0.213 & 175.1\\
   25 & 0.325 & 242.5\\
   30 & 0.392 & 269.9\\
   35 & 0.445 & 341.9\\
   40 & 0.511 & 427.9\\
   \hline
  \end{tabular}
  \caption{Computation time}
  \label{tab:compTime}
\end{table}

\subsection{Computational Complexity}
DCAD is a decentralized collision avoidance method, hence each agent makes independent decisions to avoid collision. Assuming that there are `N' agents in the environment and each agent considers `p' agents in its sensing region as its neighbors. The overall running time would be linear in the number of total agents (i.e. N). This is because the maximum number of linear constraints (ORCA velocity computation constraints) for an agent is fixed since we consider only `p' agents in its sensing region as its neighbors.

The complexity is based on Eqn. \ref{eqn:optimization} in the paper, which describes the optimization problem. The number of linear constraints due to agents dynamics, box constraints on velocity, acceleration, and jerk are fixed for an agent. The number of ORCA linear constraints for the agent depend on the number of neighboring agents considered. Since we consider only `p' in the local environment as neighbors, the maximum number of linear constraints for the optimization problem is fixed. Considering the worst case computation time to solve this optimization problem to be `k' milliseconds. The total worst case computation time for all agents in the environment would be Nk (in milliseconds) on a single core. That is a linear function of `N'.

Since the agent perform the computation independently, we can parallelize the computation on `m' cores the worst case total computation time would be kN/m.

\subsection{Benefits Over Prior Methods}
In comparison with the centralized methods shown in \cite{Preiss} and \cite{Hamer}, our method proves to be superior in handling dynamic scene changes and provides scalability to large numbers of agent due to its decentralized and reactive nature. In addition, our method is also computationally inexpensive compared to NMPC methods such as \cite{Zhu}, enabling more computation power to be available to other applications like perception and/or communication.
\section{Conclusion, Limitations, and Future Work}
In this paper, we present a decentralized collision avoidance method for a quadrotor swarm. Our method uses differential flatness and feed-forward linearization to simplify the quadrotor dynamics and uses a linear MPC to generate smooth, collision-free, downwash-conscious local trajectories.

We observe that our method results in smoother trajectories and smoother velocity variations. Further, our method performs considerably better than OCRA or AVO when following high-velocity trajectories.  We observe that our method requires 5 ms for each agent to compute a control input in the presence of 8 nearest obstacles. 

In this paper, we include a conservative method that accounts for sensor uncertainty by augmenting the bounding volume of agents and VO based on the eigenvalue of the state uncertainty. Our next step is to provide tighter bounds on the sensor uncertainty to improve the performance with noisy sensing. Further, we currently assume the position and velocity data in the case of dynamic obstacles (e.g., birds) are available through some form of sensing. It would be interesting and useful to develop robust methods with integrated sensing, tracking and planning capabilities. In addition, we plan on physically implementing our algorithm in a quadrotor system to measure its performance.
\bibliographystyle{IEEEtran} 
\bibliography{IEEEexample}
\addtolength{\textheight}{-12cm}   

\end{document}